\documentclass[letterpaper]{article} 
\usepackage{aaai2026}  
\usepackage{times}  
\usepackage{helvet}  
\usepackage{courier}  
\usepackage[hyphens]{url}  
\usepackage{graphicx} 
\usepackage{natbib}  
\usepackage{caption} 
\usepackage{algorithm}
\usepackage{algorithmic}
\usepackage{amsmath} 
\usepackage{amssymb}
\usepackage{booktabs}
\usepackage{array}
\definecolor{customgray}{gray}{0.35}

\usepackage{algorithm}
\usepackage{algorithmic}
\usepackage{multirow}
\usepackage{tabularx}
\usepackage{colortbl}
\definecolor{mygray}{gray}{.9}
\usepackage{newfloat}
\usepackage{listings}
\usepackage{newfloat}
\usepackage{listings}
\DeclareCaptionStyle{ruled}{labelfont=normalfont,labelsep=colon,strut=off} 
\floatstyle{ruled}
\newfloat{listing}{tb}{lst}{}
\floatname{listing}{Listing}
\title{Language-Guided and Motion-Aware Gait Representation for Generalizable Recognition}
\author {
    Zhengxian Wu\textsuperscript{\rm 1, †},
    Chuanrui Zhang\textsuperscript{\rm 1, †},
    Shenao Jiang\textsuperscript{\rm 1, †},
    Hangrui Xu\textsuperscript{\rm 1, \rm 2},
    Zirui Liao\textsuperscript{\rm 1},
    Luyuan Zhang\textsuperscript{\rm 1},
    Huaqiu Li\textsuperscript{\rm 1},
    Peng Jiao\textsuperscript{\rm 1},
    Haoqian Wang\textsuperscript{\rm 1, *}
}
\affiliations {
    \textsuperscript{\rm 1}The Shenzhen International Graduate School, Tsinghua University
\\
    \textsuperscript{\rm 2}School of Computer Science and Information Engineering,Hefei University of Technology\\
    {zx-wu24,zhang-cr22,jiangsa24,liao-zr24,lihq23, jiaop21}@mails.tsinghua.edu.cn, {hangruixu666,ChangYu.zly}@gmail.com \\
    \textsuperscript{†}These authors contributed equally to this work. 
    \textsuperscript{*}Corresponding author:wanghaoqian@tsinghua.edu.cn
}

\usepackage{bibentry}

\begin{document}


\maketitle

\begin{abstract}
Gait recognition is emerging as a promising technology and an innovative field within computer vision, with a wide range of applications in remote human identification. 
However, existing methods typically rely on complex architectures to directly extract features from images and apply pooling operations to obtain sequence-level representations. 
Such designs often lead to overfitting on static noise (e.g., clothing), while failing to effectively capture dynamic motion regions, such as the arms and legs. 
This bottleneck is particularly challenging in the presence of intra-class variation, where gait features of the same individual under different environmental conditions are significantly distant in the feature space.

To address the above challenges, we present a \textbf{L}anguage-guided and \textbf{M}otion-aware gait recognition framework, named \textbf{LMGait}.
To the best of our knowledge, LMGait is the first method to introduce natural language descriptions as explicit semantic priors into the gait recognition task.
In particular, we utilize designed gait-related language cues to capture key motion features in gait sequences. 
To improve cross-modal alignment, we propose the Motion Awareness Module (MAM), which refines the language features by adaptively adjusting various levels of semantic information to ensure better alignment with the visual representations.
Furthermore, we introduce the Motion Temporal Capture Module (MTCM) to enhance the discriminative capability of gait features and improve the model's motion tracking ability. 
We conducted extensive experiments across multiple datasets, and the results demonstrate the significant advantages of our proposed network. Specifically, our model achieved accuracies of 88.5\%, 97.1\%, and 97.5\% on the CCPG, SUSTech1K, and CASIAB* datasets, respectively, achieving state-of-the-art performance.
Homepage: https://dingwu1021.github.io/LMGait/
\end{abstract}


\section{1 Introduction}
Gait recognition is a biometric technology that identifies individuals based on their posture from a distance, without requiring the subject's cooperation~\cite{gaitset}. 
It has emerged as a promising modality in biometric identification. 
Compared with other methods, such as facial recognition~\cite{face-recognition} and fingerprint recognition~\cite{fingerprint-recognition}, gait recognition offers the advantages of non-contact operation and resistance to disguise. 

In gait recognition tasks, the primary goal of the model is to extract highly discriminative and robust features from complex visual inputs. 
These gait features must effectively capture the motion patterns exhibited by individuals while maintaining consistency across a variety of challenging environments. 
However, the feature extraction process in real-world scenes is often hindered by several interfering factors. 
Specifically, variations in view angle, occlusions, dynamic backgrounds and clothing changes can introduce significant changes to the visual appearance of pedestrians. 
These non-gait-related factors may dominate the visual representation, leading the network to focus on static and noisy appearance cues such as clothing rather than gait-relevant features.
This bias results in a significant increase in intra-class variations, where the gait features of the same individual under different conditions become dispersed in the feature space. 

\begin{figure}[t]
\centering
\vspace{-5pt}
\includegraphics[width=1.0\columnwidth]{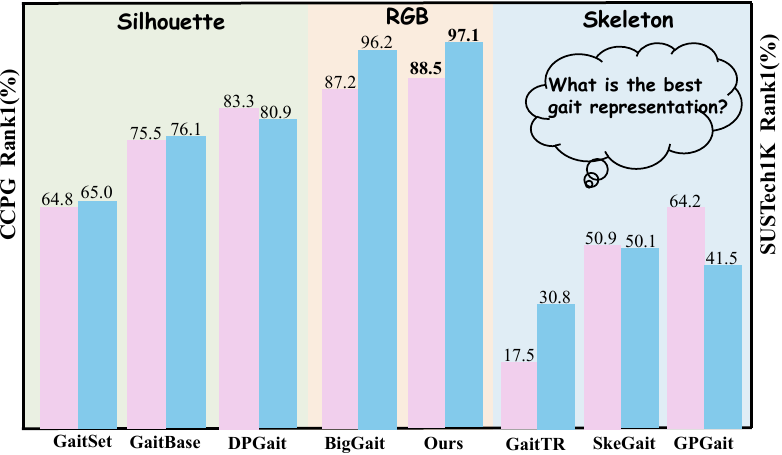} 
\caption{The Rank-1 accuracies of several networks on the CCPG~\cite{CCPG} and SUSTech1K~\cite{SUSTech1K} datasets are presented, involving three representations: Silhouette, Skeleton, and RGB. The results for CCPG are shown in red, while those for SUSTech1K are shown in blue.}
\label{fig:rank}
\end{figure}

In recent years, an increasing number of studies have focused on effectively encoding more gait-relevant cues into feature representations.
SkeletonGait++~\cite{skeletongaitgaitrecognitionusing} and ParsingGait~\cite{Parsinggait} 
leverage skeleton and human parsing maps to guide the model to focus on dynamic body parts rather than static noise. 
However, these methods are heavily dependent on precise human parsing or keypoint detection, which requires extensive manual annotations, limiting their scalability in real-world scenarios.
BigGait~\cite{BigGait} is the first to directly use raw RGB images as input, leveraging the strong feature perception ability of a pre-trained large model to extract efficient gait representation.
Nevertheless, due to the absence of an explicit mechanism to suppress irrelevant information, the introduction of more gait-relevant cues also leads to increased noise.
Moreover, although some methods enhance feature representation by designing architectures such as temporal-frequency separation~\cite{ma2023dynamic}, they rely on silhouette images, which contain less noise and are relatively easy to process. 
How to extract more gait-relevant features from the RGB input is a problem that needs to be solved.

To address the aforementioned challenges, we propose a \textbf{L}anguage-guided and \textbf{M}otion-aware gait perception framework, named \textbf{LMGait}. 
By jointly leveraging language-guided semantic priors and fine-grained temporal motion modeling, LMGait enhances dynamic gait feature extraction.
As far as we know, this is the first attempt to introduce natural language descriptions as explicit semantic priors into the gait recognition task.
Specifically, we design domain-specific textual descriptions that emphasize key body regions involved in gait. 
These semantic cues serve as informative priors and are injected into the visual feature learning pipeline to guide the model's attention toward dynamically active body parts.
 To bridge the modality gap and enhance the correspondence between textual semantics and visual features, we introduce a Motion Awareness Module (MAM), which refines language representations through learnable adapters integrated into the transformer layers.
 MAM provides a content-adaptive refinement mechanism—it learns to emphasize gait-relevant semantics while suppressing distractive cues based on the input context. 
To further enhance temporal modeling, we propose a Motion Temporal Capture Module (MTCM) to comprehensively capture part-level variations throughout the walking process. 
This module performs pixel-wise and region-level temporal feature aggregation to learn localized and continuous motion patterns.
In particular, the effectiveness of MTCM relies on prior guidance from the language branch, which facilitates the extraction of motion-related body features. 
We observe that applying temporal modeling directly on raw image features leads to a drop in accuracy, indicating that MTCM is effective only when features are concentrated on key motion regions.
As illustrated in Fig.~\ref{fig:vis1}, LMGait enables the network to focus more on motion-relevant body parts, effectively reducing interference caused by clothing variations and background noise.
\begin{figure*}[t]
\centering
\includegraphics[width=1.0\textwidth]{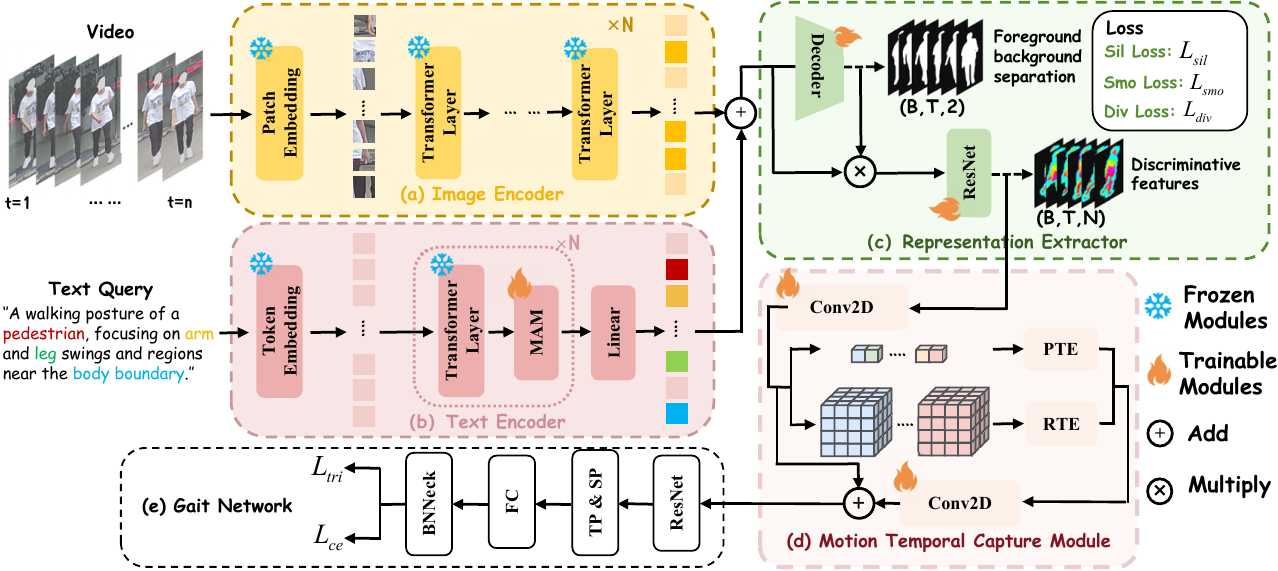} 
\caption{Pipeline of the proposed LMGait, it consists of five components.Specifically, the video input is processed through the frozen Dinov2 model for feature extraction. The text query guides the network to focus on gait-relevant regions, and it is aligned with the image feature space through the frozen CLIP text encoder and the fine-tuned MAM module. The Representation Extractor generates diverse features, while the Motion Temporal Capture Module captures posture changes during walking. Finally, the extracted features are input into the Gait Network for recognition.}
\label{baseline}
\vspace{-5pt}
\end{figure*}

We evaluate LMGait on three gait benchmarks: CCPG~\cite{CCPG}, SUSTech1K~\cite{SUSTech1K}, and CASIAB*~\cite{CASIA-B}. 
Extensive experiments demonstrate that LMGait achieves the best results in the gait recognition task. 
Notably, compared to existing methods, LMGait extracts efficient gait features, enhancing its generalization ability.
The main contributions are summarized as follows:\vspace{-0.5em}
\begin{itemize}
\item[$1)$] We leverage language cues as an explicit signal to guide the vision encoder to focus on key regions of motion. 
This approach enables our method to extract more representative gait features. 
As far as we know, this is the \textbf{first attempt} to introduce language-guided learning into the gait recognition task.
\item[$2)$] To better align textual and visual features in the latent space, we design a \textbf{Motion Awareness Module (MAM)} that dynamically adjusts the textual features.
\item[$3)$] We design a \textbf{Motion Temporal Capture Module (MTCM) } that captures the walking process of pedestrians from two complementary perspectives: the pixel-level and the region-level.
\item[$4)$]As illustrated in Fig.~\ref{fig:rank}, our method achieves state-of-the-art performance on the CCPG, SUSTech1K, and CASIAB* datasets, with Rank-1 accuracies of 88.5\%, 97.1\%, and 97.5\%, respectively.
\end{itemize}

\section{2 Related Work}
\subsection{2.1 Gait Recognition}
Gait recognition methods can be classified into two main categories: model-based and appearance-based methods. 
Model-based methods consider the basic physical structure of the human body and use interpretable models to represent gait characteristics~\cite{xu2025psgaitgaitrecognitionusing}. 
Common input modalities for these methods include 2D skeletons, meshes, and point clouds.
For example, PoseGait~\cite{posegait} leverages human pose information to extract spatio-temporal features, and employs convolutional neural networks to capture higher-level spatio-temporal representations.
DAGait~\cite{wu2025dagait} adopts a skeleton-guided data augmentation strategy, achieving improved performance in the network. 

However, model-based approaches are less effective than appearance-based methods because they lack valuable gait information such as appearance.
Appearance-based methods utilize human silhouettes to capture inherent spatial and temporal variations in body shape, clothing, and movement dynamics. 
Specifically, GaitSet~\cite{gaitset} innovatively treats the gait sequence as a set and employs a maximum function to compress the sequence of frame-level spatial features. 
GaitBase~\cite{OpenGait} leverages a universal network based on ResNet and pooling layers to produce features. 
DeepGait~\cite{deepgait} achieves promising results on real-world datasets through its deep network design. 
Biggait~\cite{BigGait} is the first approach to utilize RGB as input in an end-to-end framework for gait recognition. 
\subsection{2.2 Fine-tuning Text Encoder for Improved Visual Alignment}
While CLIP and related models perform well on general datasets, their generalization often declines on domain-specific or fine-grained tasks. 
To address this, researchers have proposed various fine-tuning strategies to improve cross-modal alignment.
One common approach is prompt tuning~\cite{prompt-tuning}, which involves using task-specific prompts to guide the text encoder toward generating embeddings that better align with visual features. 
Dual Modality Prompt Tuning (DPT)~\cite{dpt} introduces learnable prompts to both visual and text encoders, enabling joint optimization. 

In addition to prompt tuning, Low-Rank Adaptation (LoRA)~\cite{lora} offers a parameter-efficient way to enhance cross-modal alignment.
AdaLoRA \cite{adalora} further improved this approach by dynamically adjusting the rank allocation during training, thereby enabling more efficient parameter usage without compromising performance. 
Another line of work explores adapter tuning ~\cite{houlsby2019parameter}, which adds lightweight modules to frozen pre-trained models to adjust feature distributions with minimal overhead. 

\section{3 Method}
\subsection{3.1 Pipeline}
Gait recognition aims to accurately identify individuals in complex and dynamic environments.
We propose a multi-modal gait recognition framework, \textbf{LMGait}, which integrates both visual and linguistic prior knowledge to enhance the extraction of gait-discriminative features.
As illustrated in Fig.~\ref{baseline}, the input is an RGB video sequence, denoted as $X \in \mathbb{R}^{T \times H \times W \times 3}$, where $T$ represents the number of frames, and $H$ and $W$ denote the height and width of each frame.
In our setting, we randomly sample t frames from each video sequence as the model input.
The LMGait framework consists of five key components: (a)Image Encoder, (b)Text Encoder, (c)Representation Extractor, (d)Motion Temporal Capture Module and (e)Downstream gait recognition network.
For image encoding, we employ DINOv2 as the visual feature extractor.
Benefiting from its strong representation capabilities learned from large-scale pretraining, DINOv2 effectively captures the spatial distribution of human body parts.
Meanwhile, we design domain-specific textual descriptions and feed them into the Text Encoder (in Section 3.2).
To mitigate the modality gap and improve the alignment between linguistic cues and visual representations, we introduce a Motion Awareness Module (MAM).
MAM fine-tunes the Text Encoder by adaptively refining textual representations to emphasize motion-relevant semantics and suppress distractive cues.
Textual features are then fused into the visual feature map to provide semantic guidance, enabling the model to focus more effectively on motion-discriminative regions.
The Representation Extractor Module~\cite{BigGait} is designed to mitigate background interference and generate more representative feature embeddings.
To capture temporal information across frames, we design the Motion Temporal Capture Module (MTCM), which learns both pixel-level and region-level motion patterns over time.
Finally, the integrated multi-modal features are fed into the downstream gait recognition network to perform individual identity classification.
\begin{figure}[t]
\centering
\includegraphics[width=1.0\columnwidth]
{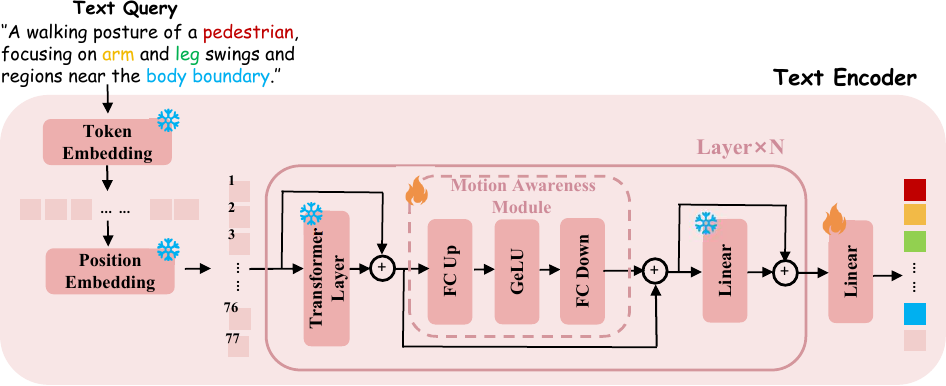} 
\caption{The architecture of the Text Encoder is designed to improve the alignment between text and image features. To achieve this, we introduce the MAM module and fine-tune it.}
\label{fig:text}
\end{figure}
\subsection{3.2 Text Encoder}
To inject domain-specific semantic priors into the visual feature learning process, we incorporate a lightweight text encoder that transforms language prompts into informative representations,  as illustrated in Fig.~\ref{fig:text}.
These textual descriptions are designed to emphasize body regions typically associated with gait motion, such as arm and leg swings, offering high-level semantic cues complementary to low-level visual features.
Formally, given an input sentence describing gait-related actions, we first tokenize the sentence using a frozen CLIP tokenizer:
\begin{equation}
\mathcal{T} = \{w_1, w_2, \ldots, w_n\}.
\end{equation}
Each token is mapped to an embedding vector $e_i$ through a lookup table:
\begin{equation}
E = [e_1, e_2, \ldots, e_n] \in \mathbb{R}^{n \times d}.
\end{equation}
We then add fixed sinusoidal positional embeddings
\begin{equation}
    P = [p_1, p_2, ..., p_n] \in \mathbb{R}^{n \times d}
\end{equation}
to incorporate sequence order, resulting in the input to the encoder:
\begin{equation}
    H^{(0)} = E + P.
\end{equation}
This embedded sequence is passed through a stack of $L$ frozen Transformer layers $\{T^{(1)}, T^{(2)}, ..., T^{(L)}\}$, each equipped with our proposed Motion Awareness Module (MAM), which adaptively adjusts the token representations:
\begin{equation}
    H^{(l)} = \text{MAM}^{(l)}\left(T^{(l)}(H^{(l-1)})\right), \quad l = 1, ..., L.
\end{equation}
Each MAM module refines the intermediate hidden states using a residual adapter structure:
\begin{equation}
    \text{MAM}(h) = h + \phi(h W_d) W_u,
\end{equation}
where $W_d \in \mathbb{R}^{d \times r}$ and $W_u \in \mathbb{R}^{r \times d}$ are trainable projection matrices with $r \ll d$, and $\phi(\cdot)$ is a non-linear activation function.

The design of MAM is based on the insight that the relevance of textual tokens varies with different visual inputs.
For example, when the lower limbs are occluded in a video frame, semantic cues related to arm movements should become more informative.
Instead of using a fixed attention mask or static prompt encoding, MAM enables the model to dynamically reweight semantic components based on their relevance to current gait scenarios. 
The final language representation is projected into a shared multimodal space using:
\begin{equation}
    z_{\text{text}} = H^{(L)} \cdot W_{\text{proj}},
\end{equation}
where $W_{\text{proj}} \in \mathbb{R}^{d \times d'}$ is a learnable matrix shared with the visual encoder's output space.

During training, the text feature $z_{\text{text}}$ is fused with the DINOv2 generated visual feature map, serving as a semantic attention mask that guides the model toward dynamically active body regions.
At the same time, to ensure efficient inference in real-world scenarios, the text encoder is used only during training to generate semantic modulation maps. 
These refined attention maps can be precomputed and cached, allowing them to be directly integrated into the visual feature stream during inference. 
As a result, the entire language branch can be removed at test time, enabling fast inference without compromising performance.
\begin{figure}[t]
\centering
\includegraphics[width=1.0\columnwidth]{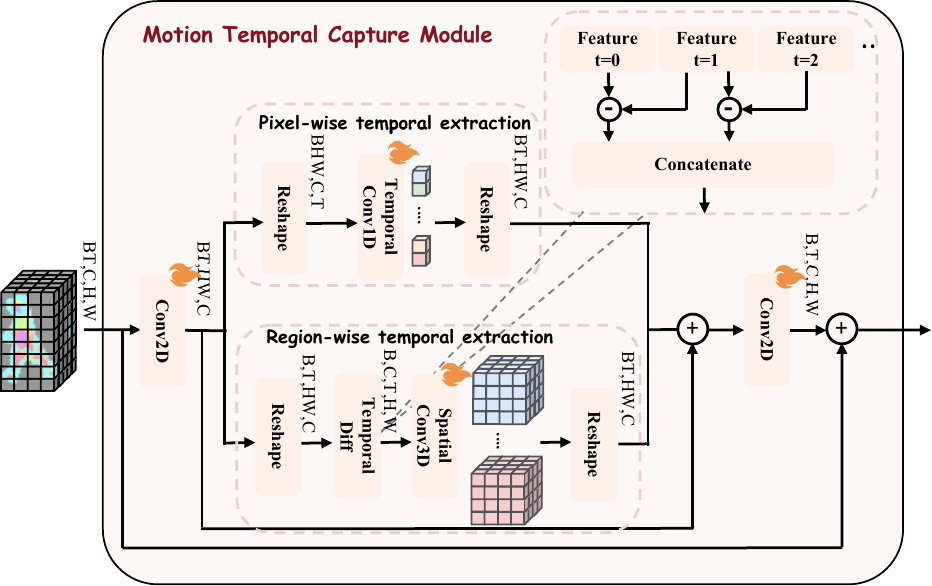} 
\caption{The architecture of the Motion Temporal Capture Module operates from two complementary perspectives: pixel-wise temporal extraction and region-wise temporal extraction. The two branches collaboratively capture gait information.}
\label{fig:time}
\end{figure}

\begin{table*}[t]
    \footnotesize
    \renewcommand{\arraystretch}{1.25}
    \centering       
    \begin{tabular}{c|c|cccccccccc}
        \toprule
            \multirow{4}{*}{Representation} & \multirow{4}{*}{Method}  & \multicolumn{10}{c}{Testing Datasets}                                                                                                                                                                     \\ \cline{3-12} 
                                     &                                                                    & \multicolumn{5}{c|}{CCPG}                             & \multicolumn{5}{c}{SUSTech1K}                     \\ \cline{3-12} 
              &  & CL   & UP   & DN   & BG   & \multicolumn{1}{c|}{Mean} & NM & CL & UF & NT & \multicolumn{1}{c}{Mean} \\ \cline{3-12}
              &                         &                 \multicolumn{5}{c|}{Rank-1}                 &          \multicolumn{5}{c}{Rank-1}      \\ \hline
            \multirow{4}{*}{skeleton} & \multicolumn{1}{c|}{GaitGraph2\cite{teepe2021gaitgraph}}    & 5.0     & 5.3       & 5.8     & 6.2    & \multicolumn{1}{c|}{5.6}    & 22.2  & 6.8    & 19.2    & 16.4                    & 18.6                    \\
             & \multicolumn{1}{c|}{Gait-TR~\cite{gaittr}}     & 15.7    & 18.3     & 18.5    & 17.5    & \multicolumn{1}{c|}{17.5}    & 33.3    & 21.0    & 34.6    & 23.5                    & 30.8                    \\
             & \multicolumn{1}{c|}{GPGait~\cite{GPgait}}     & 54.8    & 65.6     & 71.6    & 65.4    & \multicolumn{1}{c|}{64.2}    & 44.0    & 24.3    & 47.0    & 31.8                    & 41.4                    \\
             & \multicolumn{1}{c|}{SkeletonGait~\cite{skeletongaitgaitrecognitionusing}}       & 40.4      & 48.5    & 53.0   & 61.7    & \multicolumn{1}{c|}{50.9}    & 55.0    & 24.7    & 52.0    & 43.9                    & 50.1                    \\ \hline
            \multirow{4}{*}{Silhouette} & \multicolumn{1}{c|}{GaitSet~\cite{gaitset}}    & 60.2     & 65.2       & 65.1     & 68.5    & \multicolumn{1}{c|}{64.8}    & 69.1  & 61.0    & 23.0    & 65.0                    & 18.6                    \\
             & \multicolumn{1}{c|}{GaitBase~\cite{OpenGait}}     & 71.6    & 75.0     & 76.8    & 78.6    & \multicolumn{1}{c|}{75.5}    & 81.5    & 49.6    & 76.7    & 25.9      & 76.1                    \\
             & \multicolumn{1}{c|}{DeepGaitV2~\cite{deepgait}}       & 78.6      & 84.8    & 80.7   & 89.2    & \multicolumn{1}{c|}{83.3}    & 86.5    & 49.2    & 81.9    & 28.0            & 80.9                    \\ 
             & \multicolumn{1}{c|}{SkeletonGait++~\cite{skeletongaitgaitrecognitionusing}}       & 79.1      & 83.9    & 81.7   & 89.9    & \multicolumn{1}{c|}{83.7}    & 85.1    & 46.6    & 82.5    & 47.5       & 81.3                    \\ \hline
             \multirow{1}{*}{Sil+Parsing+Flow} 
        & MultiGait++~\cite{Jin2024ExploringMF} & 83.9 & 89.0 & 86.0 & 91.5 & \multicolumn{1}{c|}{87.6} & 92.0 & 50.4 & 89.1 & 45.1 & 87.4 \\ \hline
             \multirow{3}{*}{RGB} & \multicolumn{1}{c|}{GaitEdge~\cite{liang2022gaitedge}}    & 66.9     & 74.0       & 70.6     & 77.1    & \multicolumn{1}{c|}{72.7}    & -  & -    & -    & -                    & -                    \\
             & \multicolumn{1}{c|}{BigGait~\cite{BigGait}}     & 82.6    & 85.9     & 87.1    & 93.1    & \multicolumn{1}{c|}{87.2}    & 96.1    & 73.3    & 93.2    & 85.3                    & 96.2                    \\
              & \multicolumn{1}{c|}{LMGait(ours)}       & \textbf{84.8}      & \textbf{87.0}    & \textbf{88.5}   & \textbf{93.6}    & \multicolumn{1}{c|}{\textbf{88.5}}    & \textbf{96.4}    & \textbf{79.8}    & \textbf{93.9}    & \textbf{87.0}                    & \textbf{97.1}                \\         
        \bottomrule 
    \end{tabular}
    \caption{Comparison of state-of-the-art methods on the CCPG~\cite{CCPG} and SUSTech1K~\cite{SUSTech1K} datasets.}
    \label{tab1}
    \vspace{-5mm}
\end{table*}
\subsection{3.3 Motion Temporal Capture Module}
Given that human walking is a continuous and temporal process, modeling gait as a time-continuous signal is more reasonable than treating each frame independently.
To address this, we propose the \textbf{Motion Temporal Capture Module (MTCM)}, as illustrated in Figure~\ref{fig:time}.
A key novelty of MTCM lies in its synergy with the language-guided branch.
As demonstrated by our empirical analysis in Section 4.3, when RGB images are used as input, directly applying temporal modeling to unrefined visual features tends to cause the network to aggregate irrelevant or misleading cues.
Benefiting from the language-guided branch that enhances attention to motion-relevant regions, MTCM achieves more effective modeling of motion trajectories.
MTCM consists of two complementary components. 
Specifically, the \textbf{Pixel-wise Temporal Extraction (PTE)} module is designed to capture fine-grained and long-range motion patterns at individual pixel locations, while the \textbf{Region-wise Temporal Extraction (RTE)} module focuses on structured movements and posture transitions at the regional level.

\textbf{Pixel-wise Temporal Extraction (PTE)}. Let the input visual feature map sequence be 
\(\{F^{(1)}, F^{(2)}, \ldots, F^{(T)}\}\), 
where \(F^{(t)} \in \mathbb{R}^{H \times W \times C}\) is the feature map at frame \(t\). For a specific pixel location \((i,j)\), its temporal trajectory over \(T\) frames is denoted as:
\begin{equation}
x_{i,j} = [F_{i,j}^{(1)}, F_{i,j}^{(2)}, \ldots, F_{i,j}^{(T)}] \in \mathbb{R}^{T \times C}.
\end{equation}
To extract temporal dependencies at this location, we apply a 1D convolution along the temporal dimension:
\begin{equation}
y_{i,j} = \sigma(\text{Conv1D}(x_{i,j})) \in \mathbb{R}^{T' \times C'},
\end{equation}
where \(\sigma(\cdot)\) denotes a non-linear activation function.

\textbf{Region-wise Temporal Extraction (RTE)}. To capture spatial-temporal structures, RTE computes inter-frame feature differences that emphasize short-term motions:
\begin{equation}
\Delta F^{(t)} = F^{(t+1)} - F^{(t)}, \quad t = 1, \ldots, T-1
\end{equation}
The resulting sequence \(\{\Delta F^{(1)}, \ldots, \Delta F^{(T-1)}\} \in \mathbb{R}^{(T-1) \times H \times W \times C}\) captures localized motion transitions. These are stacked and passed through a 3D convolutional block:
\begin{equation}
R = \sigma(\text{Conv3D}([\Delta F^{(1)}, \ldots, \Delta F^{(T-1)}])).
\end{equation}
The outputs of PTE and RTE are fused using channel-wise concatenation followed by a 2D convolution:
\begin{equation}
F_{\text{motion}} = \text{Conv2D}([\text{PTE}(F), \text{RTE}(F)]) \in \mathbb{R}^{H \times W \times C''}.
\end{equation}
These two branches jointly enhance temporal modeling capacity.
Spefically, MTCM benefits from the text-guided visual representation. 
These semantic priors focus the input features on motion-sensitive regions , allowing MTCM to perform effective motion aggregation. 

\subsection{3.4 Representation Extractor and Loss Function}
To further enhance feature robustness, Representation Extractor aims to suppress background interference and generate diverse, discriminative feature representations.
The decoder branch is designed to reduce the feature dimensionality by producing a two-channel output, corresponding to background and foreground regions.
The silhouette supervision is only applied during training and is not required during inference.
Then, the ResNet branch is regularized by loss functions to produce smooth and discriminative features.

To achieve the best performance,we employ five loss functions: 
Triplet loss ($L_{\text{tri}}$) and cross-entropy loss ($L_{\text{ce}}$) are used to supervise the downstream gait recognition network.
\begin{equation}
L_{\text{tri}} = \left[ D(f_i, f_k) - D(f_i, f_j) + m \right]_+,
\end{equation}
\begin{equation}
L_{\text{ce}} = - \sum_{c=1}^{C} y_c \log(\hat{y}_c),
\end{equation}
$f_i$, $f_j$, and $f_k$ represent the feature embeddings of the anchor, positive, and negative samples, respectively. $m$ is the predefined margin.
Smoothness loss ($L_{\text{smo}}$) and diversity loss ($L_{\text{div}}$) are used to supervise the Representation Extractor module, encouraging the generation of smooth and diverse features.
\begin{equation}
L_{\text{smo}} = \left\| \text{Sobel}_x * \mathbf{F} \right\|_1 + \left\| \text{Sobel}_y * \mathbf{F} \right\|_1,
\end{equation}
\begin{equation}
L_{\text{div}} = - \frac{1}{C} \sum_{c=1}^{C} \sum_{i=1}^{H} \sum_{j=1}^{W} p_{c, i, j} \log(p_{c, i, j}).
\end{equation}
$\mathbf{F}$ denotes the feature map generated by the ResNet branch. $\text{Sobel}_x$ and $\text{Sobel}_y$ are the Sobel operators along the horizontal and vertical directions, respectively.$p_{c, i, j}$ is the normalized activation at position $(i, j)$ in channel $c$.
Silhouette Supervision Loss ($L_{\text{rec}}$) is employed to supervise the generation of silhouettes.
\begin{equation}
L_{\text{sil}} = - \frac{1}{HW} \sum_{i=1}^{H} \sum_{j=1}^{W} \left[ y_{i,j} \log(\hat{y}_{i,j}) + (1 - y_{i,j}) \log(1 - \hat{y}_{i,j}) \right]
\end{equation}
$\hat{y}_{i,j} \in [0, 1]$ is the predicted probability of foreground at pixel $(i, j)$. $y_{i,j} \in \{0, 1\}$ is the ground-truth binary silhouette label at pixel position $(i, j)$.
The overall loss can be formulated as:
\begin{equation}
\mathcal{L} = \mathcal{L}_{\text{tri}} + \mathcal{L}_{\text{ce}} + \alpha \mathcal{L}_{\text{sil}} + \beta \mathcal{L}_{\text{smo}} + \gamma \mathcal{L}_{\text{div}}.
\end{equation}

\begin{table}[t]
    \footnotesize
    \vspace{-3mm}
    \centering
    \begin{tabular}{c|c|cccc}
        \toprule
        \multicolumn{1}{c|}{Representation}      & \multicolumn{1}{c|}{Method}                   & NM & BG & CL  & Mean \\ \hline
        \multirow{3}{*}{Skeleton}                & \multicolumn{1}{c|}{GaitGraph2}   & 80.3 & 71.4 & 63.8 & 71.8 \\
                                                 & \multicolumn{1}{c|}{GaitTR}           & 94.7 & 89.3 & 86.7 & 90.2 \\
                                                 & \multicolumn{1}{c|}{GPGait}         & 93.6 & 80.2 & 69.3 & 81.0 \\  \hline
        \multirow{4}{*}{Silhouette}                    & \multicolumn{1}{c|}{GaitSet}       & 92.3 & 86.1 & 73.4 & 84.0 \\
                                                 & \multicolumn{1}{c|}{GaitBase}       & 96.5 & 91.5 & 78.0 & 88.7 \\
                                                 & \multicolumn{1}{c|}{GaitGCI}        & 98.4 & 96.6 & 88.5 & 94.5 \\
                                                 & \multicolumn{1}{c|}{VPNet}          & 98.3 & 96.3 & 90.7 & 94.9 \\ \hline
        \multirow{2}{*}{RGB}                     & \multicolumn{1}{c|}{BigGait}         & 100.0 & 99.6 & 90.5 & 96.7 \\
                               & \multicolumn{1}{c|}{LMGait(ours)}                       & \textbf{100.0} & \textbf{99.6} & \textbf{92.9} & \textbf{97.5} \\

        \bottomrule    
    \end{tabular}
    \caption{Comparison of Rank-1 accuracy between state-of-the-art methods on the CASIAB*~\cite{CASIA-B}.}
    \label{tab2}
    \vspace{-3mm}
\end{table}

\section{4 Experiments}
\subsection{4.1 Experimental Setup}
In this paper, our method is evaluated on three challenging gait datasets, \textbf{CCPG}~\cite{CCPG}, \textbf{SUSTech1K}~\cite{SUSTech1K} and \textbf{CASIA-B}~\cite{CASIA-B}.

The \textbf{CCPG} dataset is designed for cloth-changing gait recognition.
The dataset is split into 100 subjects for training and 100 for testing, with three cloth-changing scenarios during testing: full clothing change (CL-FULL), upper garment change (CL-UP), and lower garment change (CL-DN). 
The \textbf{CASIA-B} dataset includes 124 subjects captured from 11 different view angles, ranging from 0° to 180°. 
The \textbf{SUSTech1K} dataset encompasses various factors like visibility, view angles, occlusions and different environments.

The experimental implementation in this study followed the official protocols for each dataset~\cite{OpenGait}, including the standard partitioning strategies for the training, gallery, and probe sets. 
We adopt the average Rank-1, Rank-5, and mean Average Precision (mAP) as the evaluation metrics.
In the training stage, the number of frames for each sequence is set to 30 (for the CCPG dataset) and 24 frames (for the SUSTech and CASIAB datasets).
The Learning Rate (LR) starts at 0.1 and
is subsequently multiplied by 0.1 at the 15K, 25K, 30K, and 35K iterations. 
The SGD optimizer is adopted with a weight decay of 5e-4.
The loss weights $\alpha$, $\beta$, and $\gamma$ are 1.0, 0.01 and 5.0.
We train the model for 40k iterations with batch size of (6,8).
The number of text adapter and image transformer layers $N$ is set to 12.
The text encoder uses a vocabulary size of 49,408 and supports a maximum sequence length of 77 tokens.
To prevent the Silhouette Supervision Loss ($L_{\text{sil}}$) from dominating the overall optimization, we warm up the decoder branch with 5K iterations using only $L_{\text{sil}}$.
The decoder weights obtained from this stage are then used to initialize the decoder.
\subsection{4.2 Comparison with State-Of-The-Art Methods}
In this section, we compare the proposed LMGait with several popular state-of-the-art(SOTA) gait recognition methods.
The experimental results on the CCPG and SUSTech1K datasets are presented in Table \ref{tab1}.
The results on the CASIAB* dataset are shown in Table \ref{tab2}.
LMGait achieves significantly better performance than other SOTA methods.

Taking the CCPG dataset as an example, LMGait outperforms the current skeleton-based SOTA method GPGait by 24.3\% in Rank-1 accuracy (88.5\% vs. 64.2\%).
For silhouette-based methods, which are the most widely used representations in gait recognition, LMGait surpasses the SOTA method SkeletonGait++ by 4.8\% (88.5\% vs. 83.7\%). 
Using RGB images directly as input poses greater challenges for focusing on gait-related cues.
Currently, GaitEdge and BigGait are the only two methods that operate directly on RGB inputs. 
Compared to BigGait, our method achieves improvements of 2.2\%, 1.1\%, 1.4\%, and 0.5\% in the CL, UP, DN, and BG scenarios, with an average gain of 1.3\% (88.5\% vs. 87.2\%).
We then conducted cross-dataset evaluation experiments by training our network on the CCPG dataset and evaluating it on the SUSTech1K and CASIA-B* datasets.
The results are presented in Table \ref{tab3}.
These datasets exhibit significant variations in terms of subjects and scenes. 
Compared to other methods, LMGait, by focusing on discriminative information, achieves higher accuracy in cross-dataset evaluations. 
We provide the complete evaluation results on all three datasets, as well as additional cross-dataset evaluation results, in the supplementary material.
Based on the above analysis, we conclude that LMGait extracts better gait features and mitigates the interference from irrelevant noise.
\begin{table}[t]
    \footnotesize
    \vspace{-3mm}
    \centering
    \begin{tabular}{c|ccc|ccc} 
    \hline
    \multirow{3}{*}{Model} & \multicolumn{6}{c}{Test Set}                                   \\ 
    \cline{2-7}
                           & \multicolumn{3}{c|}{CASIA-B*} & \multicolumn{3}{c}{SUSTech1K}  \\
                           & NM   & BG   & CL              & Clothing & Night & Overall     \\ 
    \hline
    GaitSet                & 47.4 & 40.9 & 25.8            & 8.2      & 11.0  & 12.8        \\
    GaitBase              & 59.1 & 52.7 & 30.4            & 9.5      & 13.1  & 16.8        \\
    BigGait                 & 77.4 & 71.5 & 33.6            & 43.7     & 44.8  & 56.4        \\
    Ours                 & \textbf{81.6} & \textbf{75.4} & \textbf{35.4}            & \textbf{45.6}     & \textbf{48.2} & \textbf{59.3}        \\
    \hline
    \end{tabular}
    \caption{Cross-dataset evaluation is conducted by training the model on CCPG and testing on CASIA-B* and SUSTech1K with Rank-1 accuracy as the evaluation metric.}
    \label{tab3}
\end{table}


\begin{table}[t]
    \footnotesize
    \vspace{-3mm}
    \centering
    \begin{tabular}{c|cccc}
        \toprule
        Method                  & CL & UP  & DN & Mean\\ \hline
        w/o Text Branch \& w/o MTCM     & 82.6      & 85.9    & 87.1 & 87.2\\
        w/o Text Branch \& w/ 3D-CNN         & 82.3      & 85.0    & 86.4   & 86.4\\
        w/o Text Branch \& w/ MTCM     & 82.8      & 84.5    & 87.2   & 86.9\\
        w/ Text Branch \& w/ MTCM                      & \textbf{84.8}      & \textbf{87.0}    & \textbf{88.5}  & \textbf{88.5}\\
        \bottomrule    
    \end{tabular}
    \caption{Ablation study on the correlation between the Text Branch and the Motion Temporal Capture Module (MTCM).}
    \label{tab4}
    \vspace{-5mm}
\end{table}

\begin{table}[t]
    \footnotesize
    \vspace{-3mm}
    \centering
    \begin{tabular}{c|ccccc}
        \toprule
        Method                  & CL & UP  & DN & BG & Mean\\ \hline
        w.o. PTE-TPConv1D     & 82.9      & 86.2    & 87.6 & 93.2 & 87.5\\
        w.o. RTE Temp-Diff         & 83.8      & 86.8    & 88.5  & 93.5 & 88.2\\
        w.o. RTE-SPConv3D     & 83.6      & 86.6    & 88.2   & 93.4 & 87.9\\
        Ours                       & \textbf{84.8}      & \textbf{87.0}    & \textbf{88.5}  & \textbf{93.6} & \textbf{88.5}\\
        \bottomrule    
    \end{tabular}
    \caption{The contribution of different components within the MTCM Module to Rank-1 Accuracy(on the CCPG dataset).}
    \label{tab5}
    \vspace{-3mm}
\end{table}

\begin{table}[t]
    \footnotesize
    \centering
    \begin{tabular}{c|cccc}
        \toprule
        Method                  & CL & UP  & DN & Mean\\ \hline
        w.o.\ Text Branch \& MAM     & 82.8      & 84.5    & 87.2 & 86.9\\
        w.o.\ MAM + Wrong Text      & 81.6      & 83.8    & 86.8  & 86.1\\
        w. MAM + Wrong Text    & 82.8      & 84.9    & 87.7   & 87.1\\
        w.o.\ MAM + Correct Text     & 83.1      & 86.4    & 87.5   & 87.6\\
        w. MAM + Random Text    & 83.8      & 86.6    & 88.2   & 88.1\\
        w.MAM \& Correct Text                       & \textbf{84.8}      & \textbf{87.0}    & \textbf{88.5}  & \textbf{88.5}\\
        \bottomrule    
    \end{tabular}
    \caption{Effect of the Text Encoder on Rank-1 Recognition Accuracy(on the CCPG dataset).}
    \label{tab6}
    \vspace{-3mm}
\end{table}

\begin{table}[t]
    \footnotesize
    \vspace{-3mm}
    \centering
    \begin{tabular}{c|ccccc}
        \toprule
        Visual Encoder                  & CL & UP  & DN & BG & Mean\\ \hline
        CLIP Image Encoder     & 68.7      & 75.1    & 72.2 & 79.4 & 73.9\\
        ours                       & \textbf{84.8}      & \textbf{87.0}    & \textbf{88.5}  & \textbf{93.6} & \textbf{88.5}\\
        \bottomrule    
    \end{tabular}
    \caption{The impact of different visual encoders on accuracy.}
    \label{tab7}
    \vspace{-3mm}
\end{table}

\begin{figure}[t]
    \centering
    \includegraphics[width=1\linewidth]{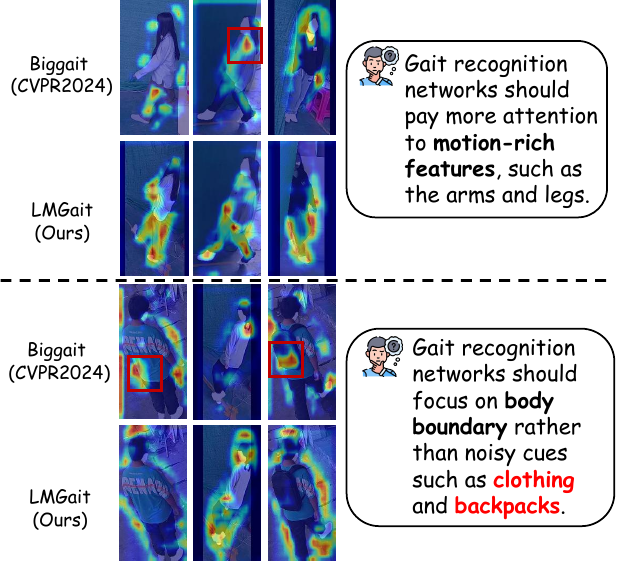}
    \vspace{-1em}
    \caption{Visualization of attention maps from the first layer of the downstream gait recognition network.}
    \label{fig:vis1}
    \vspace{-3mm}
\end{figure}

\subsection{4.3 Ablation Study and Visualization}
\textbf{The correlation between the Motion Temporal Capture Module (MTCM) and the Text Encoder.} 
The MTCM and the text branch serve complementary roles in our framework. 
Under the explicit guidance of textual priors, the network is encouraged to focus on motion-relevant regions—such as the arms and legs—which enables MTCM to more effectively capture temporal motion patterns. 
As shown in Table~\ref{tab4}, when the text branch is removed, using either MTCM or an alternative temporal module (3D-CNN) results in a performance drop compared to the baseline without any temporal modeling (second and third row vs first row).
In contrast, the full model (bottom row), which incorporates both the text branch and MTCM, achieves the highest accuracy across all scenarios.

\textbf{Motion Temporal Capture Module(MTCM).}
The MTCM module consists of two complementary branches: pixel-wise Temporal Extraction (PTE) and region-wise Temporal Extraction (RTE). 
These branches work collaboratively to enhance the model’s capability in capturing temporal dynamics.
As shown in Table~\ref{tab5}, removing any component of MTCM leads to a drop in overall performance.

\textbf{Text Encoder.}
We further evaluate the contribution of the text branch to the overall performance, as summarized in Table~\ref{tab6}. 
When the text branch is removed, a substantial drop in accuracy is observed (86.9\% vs. 88.5\%). 
This performance drop is caused by the MTCM module aggregating information that is irrelevant to gait.
This issue becomes even more pronounced when incorrect textual inputs are provided to the model. 
Specifically, we constructed a misleading text query: \textbf{''Attention is given to environmental details such as the blue sky, the clean ground, and the clear windows.''}
When this erroneous information is integrated into the feature maps without any refinement, the accuracy drops significantly by 2.4\% (86.1\% vs. 88.5\%).
Introducing a refinement module for these incorrect texts helps mitigate such adverse effects to some extent. 
When correct text descriptions are used but the refinement module is disabled, the network still benefits from the semantic priors provided by language encoder, guiding attention toward gait-relevant body. 
Meanwhile, we deployed the MLLM~\cite{Li2024LLaVAOneVisionEV} to generate generic descriptions.
After six test runs, the average performance was lower than that achieved with our manually designed text inputs (88.1 \% vs 88.5\%). Detailed information can be found in our supplementary material.

\textbf{Visual Encoder and Inference Efficiency}.
Table~\ref{tab7} presents a comparison between two visual encoders, CLIP and DINOv2, under the same framework. 
While the image branch of CLIP may be better aligned with the text branch in the shared feature space, DINOv2 is better at capturing fine-grained visual features.
In the supplementary material, we also analyze that fine-tuning the visual encoder can accelerate the convergence process but leads to a drop in accuracy.
 In terms of \textbf{inference efficiency}, gait recognition often demands real-time performance in applications. 
 In our framework, the text encoder is only used during training to generate semantic attention guidance related to gait. 
 Once training is complete, the resulting attention maps can be stored and directly integrated with visual features at inference time, eliminating the need to run the text branch again. 
Removing the \textbf{loss} supervision from the Representation Extractor also leads to a performance drop.Detailed ablation studies on the visual encoder, inference efficiency, representation extractor, and loss functions can be found in our supplementary material.
As shown in Figure~\ref{fig:vis1}, we visualize the attention maps from the first layer of the downstream gait recognition network. 
In contrast, our method effectively extracts more discriminative features from gait-relevant regions.
\section{5 Conclusion}
In this paper, we present a language-guided and motion-aware gait recognition framework, named LMGait. 
This is the first attempt to introduce natural language descriptions as explicit semantic priors into the gait recognition task. 
With the guidance of semantic cues, the network can extract features that are more relevant to gait.
To further align the text and image spaces, we design the Motion Awareness Module (MAM) to fine-tune the textual features. 
Furthermore, we propose the Motion Temporal Capture Module (MTCM) to capture the changes in posture as a person walks.
The text branch and the MTCM work collaboratively to focus the features on motion-related body regions.
Results show LMGait outperforms previous methods on multiple datasets.

\section{6 Acknowledgments}
This work is supported by the Shenzhen Science and Technology Project under Grant KJZD20240903103210014.

\bibliography{aaai2026}


\end{document}